\pdfoutput=1

\documentclass[11pt]{article}

\usepackage[preprint]{acl}

\usepackage{times}
\usepackage{latexsym}
\usepackage{booktabs}
\usepackage[T1]{fontenc}

\usepackage{mdframed}
\usepackage{amssymb}

\usepackage{fancyvrb}
\usepackage[T1]{fontenc}
\usepackage[utf8]{inputenc}
\usepackage{microtype}
\usepackage{multirow}
\usepackage{amsmath}
\usepackage{cleveref}
\usepackage{enumitem}
\usepackage{fontawesome5}
\usepackage{inconsolata}
\usepackage{array}
\usepackage{graphicx}

\title{Benchmarking Deep Search over Heterogeneous Enterprise Data}

\author{\quad Prafulla Kumar Choubey \quad Xiangyu Peng \quad Shilpa Bhagavath \\
 {\bf Kung-Hsiang Huang \quad Caiming Xiong  \quad Chien-Sheng Wu}\\
Salesforce AI Research \\
\small{
   \textbf{Correspondence:} \href{mailto:pchoubey@salesforce.com,wu.jason@salesforce.com}{[pchoubey, wu.jason]@salesforce.com}} \\
}

\iftrue

\NewDocumentCommand{\steeve}
{ mO{} }{\textcolor{brown}{\textsuperscript{\textit{Steeve}}\textsf{\textbf{\small[#1]}}}}

\else
\newcommand{\steeve}[1]{}

\fi

\newcommand{\datasetname}{HERB}

\begin{document}
\maketitle
\begin{abstract}

We present a new benchmark for evaluating Deep Search—a realistic and complex form of retrieval-augmented generation (RAG) that requires source-aware, multi-hop reasoning over diverse, sparsed, but related sources. These include documents, meeting transcripts, Slack messages, GitHub, and URLs, which vary in structure and often contain human-to-human interactions.
We build it using a synthetic data pipeline that simulates business workflows across product planning, development, and support stages, generating interconnected content with realistic noise and multi-hop questions with guaranteed ground-truth answers.
We release our benchmark with both answerable and unanswerable queries, and retrieval pool of 39,190 enterprise artifacts, enabling fine-grained evaluation of long-context LLM and RAG systems. 
Our experiments reveal that even the best-performing agentic RAG methods achieve an average performance score of 32.96 on our benchmark. With further analysis, we highlight retrieval as the main bottleneck: existing methods struggle to conduct deep searches and retrieve all necessary evidence. Consequently, they often reason over partial context, leading to significant performance degradation~\footnote{\scriptsize{
\begin{tabular}{@{}l@{}}
\ \faGithub\ \ \url{https://github.com/SalesforceAIResearch/HERB} \\
\raisebox{-0.5ex}{\includegraphics[height=3ex]{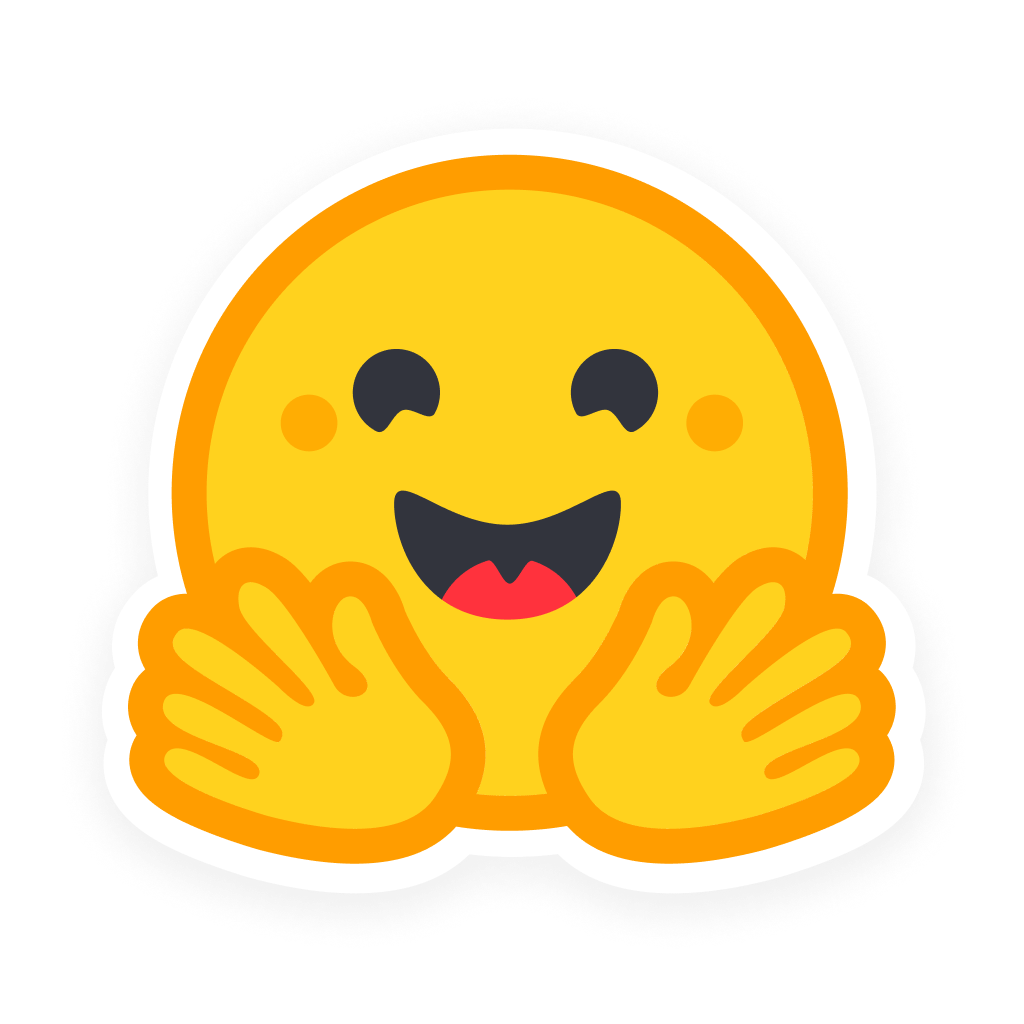}}\ \url{https://huggingface.co/datasets/Salesforce/HERB}
\end{tabular}}}.

\end{abstract}

\section{Introduction}
RAG has gained widespread adoption in enterprise applications for tasks that require grounded responses ~\cite{packowski2024optimizingevaluatingenterpriseretrievalaugmented,cohen2025wixqamultidatasetbenchmarkenterprise,yu-etal-2025-ekrag}. However, current multi-hop RAG benchmarks typically build questions over clusters of related documents, connected through explicit entity or topical links inferred by LLMs~\cite{ragas2024}. 
This approach makes weak connection between text chunks, and the questions generated on top of these chunks can be artificial and superficial. 
For example, consider a query from the MultiHop-RAG benchmark~\cite{tang2024multihoprag}: \textit{``Which platform is at the center of discussions in articles from Music Business Worldwide, Polygon, and FOX News - Health, concerning the policing of AI-driven voice replication, the debate over "reaction" content, and being the most used app overnight by young people?''} 
The question is overly specific and unnatural, and does not reflect real-world use cases. These benchmarks also involve limited reasoning using shallow search, allowing RAG systems to achieve high scores.


To better reflect the challenges of real-world Deep Search tasks, we introduce \textbf{{\datasetname}}, a \textbf{H}eterogeneous \textbf{E}nterprise \textbf{R}AG \textbf{B}enchmark inspired by common use cases in the software industry. Deep Search is a retrieval-centric task that requires not only determining what information to search for but also where to search for it—often relying on real-world knowledge to navigate heterogeneous sources and identify the most relevant context. 
For instance, determining who provided feedbacks to a document may seem straightforward using structured document metadata (e.g., comment or edit history). 
However, feedbacks can be provided across unstructured sources like Slack messages or meeting, which are only surfaced through a deeper search. 
We define Deep Search as distinct from Deep Research \cite{shao2024assisting,openaideepresearch}, the latter involves conducting multi-step research on a topic, often using web browsing, data analysis, coding, and report generation.

\begin{figure*}[tb!]
\centering
\includegraphics[width=0.98\linewidth]{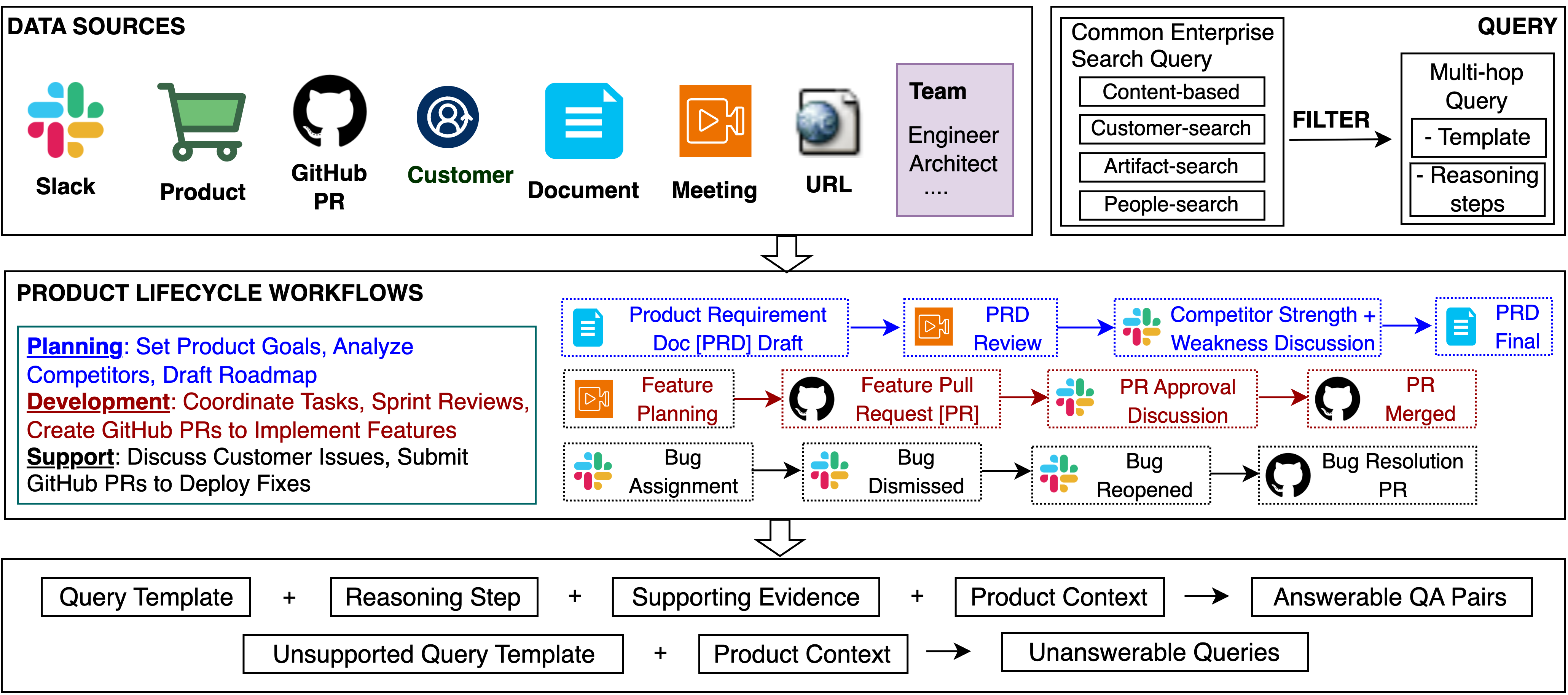}
\caption{Overview of the workflow-guided data synthesis framework. Team roles(e.g., Architect) engage through meetings and slack channels. Interactions generate artifacts (e.g., meeting transcripts, github PRs, documents) which form the basis for query generation.}
\label{fig:fig1}
\end{figure*}

We build {\datasetname} by first collecting common enterprise queries on content, people, artifact, and customer. 
Then, to generate the contextual data needed to answer these queries, we use LLMs to simulate realistic enterprise environments. Inspired by CRMArena~\cite{huang-etal-2025-crmarena,huang2025crmarena}, which demonstrated the effectiveness of LLMs for scalable enterprise data generation, we construct a representative enterprise scenario centered on three stages of the software product lifecycle: planning, development, and deployment (\Cref{fig:fig1}). The planning stage focuses on defining goals and roadmaps, the development stage centers on implementing features and coordinating engineering tasks, and the deployment stage involves resolving customer issues and shipping fixes.  For each stage, we design three realistic workflows that reflect typical enterprise processes, enabling the generation of rich contextual data. For example, a development-stage workflow may involve teams coordinating through meetings, assigning engineering tasks, tracking progress via Slack, and managing implementation using GitHub—closely mirroring real-world software development practices.

Using our pipeline, we generate a comprehensive synthetic dataset that reflects the scale and heterogeneity of a small-scale software organization. Our simulated enterprise includes 530 employees working across 30 products. For each product, we simulate three-stage workflows (planning, development, and deployment), resulting in a total of 39,190 data artifacts spanning structured and unstructured sources. We generate 815 answerable queries, each supported by specific evidence found within a subset of relevant artifacts. Additionally, to support robust evaluation of model precision and failure modes, {\datasetname} also includes 699 unanswerable queries, inspired by~\citet{peng2025unanswerabilityevaluationretrievalaugmented}, by mapping realistic questions to workflows that lack any supporting evidence.

We evaluate a range of RAG systems and LLM configurations on {\datasetname} to assess their ability to handle complex, enterprise-style queries. 
Our results show that standard RAG methods—whether using linear chunking or graph-based retrieval—struggle with the complexity of enterprise data. The best-performing baseline, a hybrid of dense and sparse retriever, achieves only an average performance of 20.61.
Agentic RAG systems with planning and tool-use capabilities perform best overall but still achieve only a modest average performance of 32.96, highlighting significant room for improvement.

{\datasetname} also supports a secondary evaluation setting focused on the long-context reasoning abilities of LLMs. In this setup, we provide the model with the full set of product-specific artifacts—instead of retrieved passages—and task it with answering questions by identifying and reasoning over relevant information in-context. We find that all models struggle in this setting as well. For example, Gemini-2.5-Flash, the best-performing model, achieves an average performance of only 76.55. Older or weaker models, such as Llama-3.1-405B-Instruct, perform significantly worse, with an average performance of just 18.20. 

Designed to stress-test RAG systems, {\datasetname} captures complex, dispersed knowledge across heterogeneous sources, features natural multi-hop questions requiring cross-source reasoning, and emphasizes realistic, information-seeking queries aligned with enterprise needs—addressing key limitations of prior datasets.


\section{Related Work}
\paragraph{Multihop RAG Evaluation:} Multihop RAG benchmarks typically follow a post-hoc approach, where they first form clusters of related documents or claims and then generate questions over the linked content. For example, HotpotQA~\cite{yang-etal-2018-hotpotqa}, 2WikiMultihopQA~\cite{ho-etal-2020-constructing}, MuSiQue~\cite{trivedi-etal-2022-musique}, and MultiHopRAG~\cite{tang2024multihoprag} created multihop questions by linking claims through bridging entities. Recently, MHTS~\cite{lee2025mhtsmultihoptreestructure} replaced entity-based linking with clustering, grouping semantically similar claims to form multihop queries. RAGAS~\cite{ragas2024} presents a general framework that connects claims or documents using various similarity signals, such as entity overlap, topical similarity or shared keywords. 

\paragraph{Query-First Synthesis of Evaluation Data:}
SummHay~\cite{laban-etal-2024-summary} presented a query-first design, generating documents conditioned on predefined topics and insights. {\datasetname} builds on this approach but introduces two key innovations. First, whereas SummHay emphasizes shallow reasoning over explicit insights, {\datasetname} targets deeper inductive reasoning over implicit information. Second, SummHay operates exclusively on a single unstructured source, while {\datasetname} integrates both structured data and unstructured text, enabling cross-format reasoning. To the best of our knowledge, no existing RAG benchmark supports implicit reasoning across heterogeneous data formats.

\paragraph{Synthetic Enterprise Environment:}
Recent work~\cite{drouin2024workarenacapablewebagents,styles2024workbenchbenchmarkdatasetagents,yao2024taubenchbenchmarktoolagentuserinteraction,drouin2024workarenacapablewebagents,boisvert2024workarena++,xu2025theagentcompanybenchmarkingllmagents,huang-etal-2025-crmarena,huang2025crmarena} has introduced synthetic enterprise environments to evaluate LLM agents on task execution. 
In contrast, our work focuses on answering complex multihop queries that require agents to model and reason  about enterprise processes. Instead of completing tasks directly, agents must identify which data sources contain relevant information, determine what to retrieve, and integrate evidence across heterogeneous formats to produce a correct answer.

\begin{table*}[ht]
\small
\centering
\resizebox{0.95\linewidth}{!}{%
\begin{tabular}{p{0.11\linewidth} p{0.07\linewidth} p{0.24\linewidth} p{0.64\linewidth}}
\toprule
\textbf{Stage} & \textbf{Type} & \textbf{Query} & \textbf{Reasoning Scenario} \\
\midrule
Planning & People & Find employee ID of the author and reviewers of the \textit{\{product\}}'s PRD? & Locate the Product Requirements Document (PRD) for \textit{\{product\}} $\rightarrow$ 
Identify the author of the PRD $\rightarrow$ 
Trace related evidence across slack and meeting transcripts $\rightarrow$ 
Identify individuals who provided feedback $\rightarrow$ 
Link all identified individuals to employee IDs
\\
\midrule
Development & Content & What features for \textit{\{product\}} were discussed but not implemented? & Extract proposed features from Slack threads and meeting transcripts/ chats $\rightarrow$ Map features to implementation records in GitHub PRs $\rightarrow$ Identify features without corresponding implementations \\
\midrule
Support & Customer & Find the name of company that has the the highest number of unresolved bugs in \textit{\{product\}}? & Identify all unresolved bugs for \textit{\{product\}} (i.e., bugs without a linked PR) $\rightarrow$ 
Associate each bug with the relevant customer using Slack discussions or meeting transcripts $\rightarrow$ 
Aggregate counts to find the customer with the most open issues \\
\midrule
All & Artifact & Find the demo URLs shared by team members for \textit{\{product\}}'s competitor products? & Search Slack and meeting transcripts for competitor product names $\rightarrow$ For each identified competitor product, search Slack and meeting transcripts again to extract any shared demo URLs \\ \bottomrule
\end{tabular}}
\caption{Examples of queries and the reasoning required to answer them using structured and unstructured data.
}
\label{tab:serb-query-reasoning}
\end{table*}

\section{Data Generation Pipeline}


To address the 
challenges with
post-hoc synthetic RAG evaluations, {\datasetname} adopts a query-first data synthesis strategy. It begins by manually defining realistic enterprise queries 
and then synthesizing the supporting evidence required to answer them. 
We first describe the synthetic enterprise setup (\S\ref{sec:data-setup}), followed by our query selection methodology (\S\ref{sec:data-query}), the workflow-based data generation process (\S\ref{sec:data-workflow}), and the final dataset statistics (\S\ref{sec:data-stats}). 


\subsection{Data Schema Setup} \label{sec:data-setup}

We construct an enterprise schema that reflects the structure of a simplified modern software company. The environment comprises six functional organizations (Slack, Sales, Einstein, Salesforce, Tableau, and Mulesoft), each managing five fictional products with distinct descriptions. 
Each organization has 10 cross-functional roles, such as software engineers and product managers, with one or more employees assigned to each role per team. 
Every employee is assigned a unique employee ID, while real-world ambiguity is preserved by allowing duplicate names for different individuals.
To incorporate customer-facing interactions, we also generate customer profiles, each of which contains a named contact, a unique customer ID, a company name, and a list of associated products.


Each employee interacts across multiple Slack channels, and each channel is corresponding to a specific phase of the product lifecycle.
For example, product planning channels include product managers, marketing managers, and engineering leads, while development channels involve software engineers and other relevant roles. 
Employees also participate in meetings, each of which is associated with a transcript and may include a chat log if participants share additional information, such as URLs. 
In addition, employees author and review documents, create and review pull requests (PRs), and share documents, PRs, and URLs through Slack messages or during meetings.
This structure mirrors real-world enterprise settings. 
Examples of different data objects are shown in \Cref{app:schema-objects}. 



\subsection{Realistic Query Selection} \label{sec:data-query}
We focus on four prevalent enterprise search intents: (1) content-based queries (e.g., employee activities or product feedback), (2) people-search queries (e.g., internal employee information), (3) artifact-search queries (e.g., GitHub pull requests, documents, or URLs), and (4) customer-search queries (e.g., customer-related information). From these categories, we focus on complex queries that require multi-hop reasoning. For each selected query, we construct one or more corresponding reasoning scenarios grounded in real-world enterprise contexts, explicitly outlining the inference steps needed for resolution. ~\Cref{tab:serb-query-reasoning} presents example queries for each content type. 

To support precise evaluation, we avoid under-specified questions that admit multiple plausible answers. For example, instead of the ambiguous formulation ``Which customer has the highest number of unresolved bugs in \textit{\{product\}}?'' we use a deterministic alternative: ``Find the name of the company with the highest number of unresolved bugs in \textit{\{product\}}.'' This explicitly constrains the expected answer format to the company name, avoiding ambiguity between possible alternatives such as customer ID, contact person, or company name. 
All query selection and reasoning scenario design steps are performed manually by domain experts to ensure clarity and alignment with enterprise search tasks. 

In total, we define 41 query templates (see Appendix \Cref{tab:question-templates}): 12 for content-based queries, 14 for people-search, 8 for artifact-search, and 7 for customer-search. We then use these templates to generate specific queries conditioned on product context and other metadata such as employee roles or document types. 

\begin{figure*}[tb!]
\centering
\includegraphics[width=0.98\linewidth]{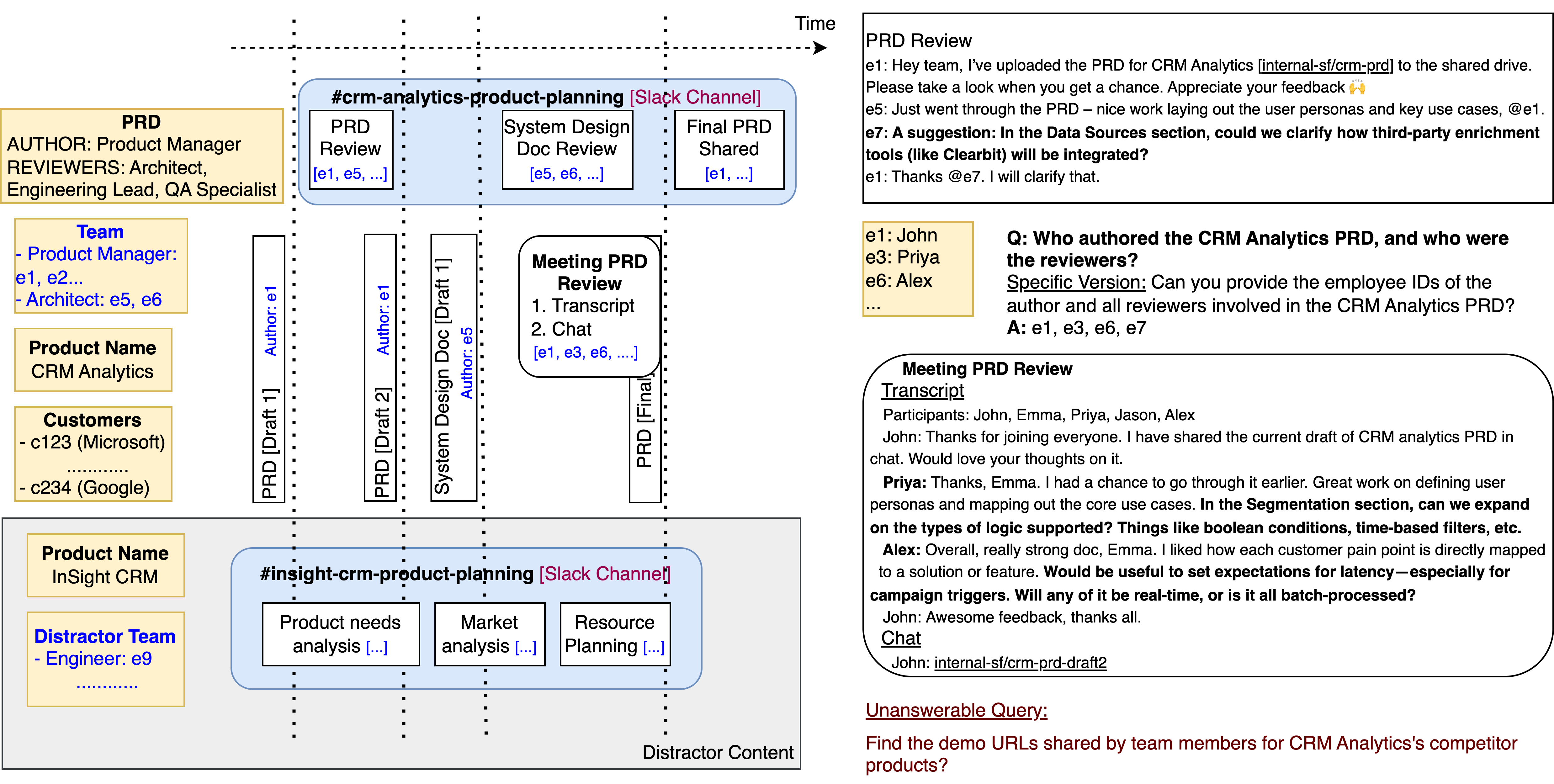}
\caption{Illustration of a product-planning workflow from the {\datasetname} dataset. A product manager shares a product requirements document (PRD) for review via Slack, updates it based on feedback, and then presents a revised draft in a meeting. The team provides additional input, leading to further iterations. These interactions and artifacts form the basis for query generation in {\datasetname}.}
\label{fig:stage_1}
\end{figure*}

\subsection{Human-Designed Workflows} \label{sec:data-workflow}

After defining realistic queries, we generate supporting evidence using manually designed workflows aligned with the reasoning required for each query. These workflows simulate how information is produced and shared in enterprise settings, yielding artifacts such as Slack discussions and meeting transcripts. Some artifacts contribute directly to answering a query, while others serve as distractors, reflecting the ambiguity and noise typical of real-world scenarios.
Workflows are organized around three core stages of the product lifecycle:

\begin{itemize}[leftmargin=*]
\item \underline{Product Planning}: Teams set goals, analyze competitors, and draft roadmaps through documents, slack discussions, and planning meetings.
\item \underline{Product Development}: Teams coordinate tasks, review sprints, and create GitHub PRs to implement features.
\item \underline{Product Support}: Teams discuss customer issues, assign owners, track progress, and submit PRs to deploy fixes.
\end{itemize}

Each query is linked to one or more consecutive stages and workflows based on its reasoning requirements. See~\cref{tab:serb-query-reasoning} for examples, where each query is paired with one possible reasoning path. 

For each stage, we define three workflows 
capturing different enterprise activity patterns, and group queries with similar reasoning needs within the same workflow. We provide an overview of all nine workflows in Appendix~\Cref{app-tab:serb-wfs}. To discourage overfitting, full workflow specifications are not released publicly but are available upon request for research purposes. 

\paragraph{An Example Workflow:}
In~\autoref{fig:stage_1}, we illustrate an example Stage 1 (Product Planning) workflow. A common people-centric query at this stage involves identifying contributors to key planning artifacts, such as the Product Requirements Document (PRD). In enterprise settings, these documents are typically drafted and refined through iterative collaboration across Slack discussions and meetings. 

The process begins by selecting the product metadata, relevant team members, and associated customers, which together provide shared context across all three stages. It then orchestrates a sequence of collaborative activities that reflect typical enterprise behavior. For example, in~\autoref{fig:stage_1}, employee \texttt{e1} drafts the PRD and shares it in Slack, where \texttt{e7} provides feedback. \texttt{e1} then revises the document and presents it in a planning meeting, where \texttt{e3} and \texttt{e6} offer additional suggestions. The meeting transcript includes only participant names, introducing ambiguity consistent with real-world records. A separate structured source maps unique employee IDs to their names, enabling entity disambiguation. After incorporating the feedback, \texttt{e1} finalizes the PRD. All generated data are timestamped to 
enable RAG systems to leverage time-based reasoning when needed.

\paragraph{Query Grounding and Answer Generation:} As each workflow is executed, artifacts are explicitly linked to the queries they support to ensure traceability between questions and evidence. 
For example, in~\autoref{fig:stage_1}, Slack messages, PRD documents, and meeting transcripts related to the PRD review are linked to the query: ``Can you provide the employee IDs of the author and all reviewers involved in the CRM analytics PRD?'' 

To construct an answer, we do not directly use an LLM to generate responses given linked evidence, as this can be unreliable for multi-hop questions. Instead, we break the task into interpretable steps. We first aggregate all artifacts associated with a query and then apply a structured, query-specific inference process. For the example question above, we 1) identify utterances in the linked Slack messages and meeting transcript that contain review-related feedback, and 2) the authors of the PRD documents. We then resolve the corresponding speakers and authors by mapping their names to employee IDs to obtain the final answer. 

\paragraph{Unanswerability Evaluation}
Motivated by UAEval4RAG~\cite{peng2025unanswerabilityevaluationretrievalaugmented}, {\datasetname} also includes unanswerable queries to evaluate a RAG system’s ability to recognize when information is missing. These queries are drawn from the full set of query templates but are paired with workflows that lack the required supporting evidence. For example, since the workflow in~\Cref{fig:stage_1} do not contain discussion about competitor products, we define unanswerable queries such as: ``Find the demo URLs shared by team members for CRM Analytics's competitor products.''

\paragraph{Realistic Noise in Workflows-Generated Data}
Our data naturally includes noise, as enterprise workflows often involve overlapping activities and partial information. 
For example, in~\autoref{fig:stage_1}, while the PRD review is underway, \texttt{e5} simultaneously authors a system design document and shares it in Slack for feedback. These concurrent actions distribute relevant evidence across Slack messages and meeting transcripts, requiring the model to aggregate and filter context from multiple sources.

To further increase task complexity, we introduce distractor content that mirrors the noise commonly found in real-world enterprise settings. 
For instance, multiple teams may appear to contribute to the same product during the planning stage, although only one continues development in later phases. 
Additionally, beyond discussing their own pull requests (PRs), teams may reference PRs from open-source libraries or other internal projects. We list the full set of distractor types in ~\Cref{app:distractor}.

\subsection{Dataset Statistics} \label{sec:data-stats}
In total, the dataset includes 530 employees distributed across three teams per organization, covering 30 products. It contains 302 Slack channels with 33,632 messages, 400 documents, 3,562 pull requests, 575 shared URLs, 321 meeting transcripts, 50 meeting chats, and 120 customer profiles. The dataset comprises 815 answerable queries, including 238 content-based, 260 people-search, 130 customer-search, and 187 artifact-search queries. In addition, it includes 699 unanswerable queries. 
While the dataset is primarily designed for evaluating RAG systems, it also supports long-context reasoning, where LLMs are provided with all product-related content and must reason over large, structured and unstructured contexts to answer complex queries.

\section{Experiments}

\subsection{Experimental Settings}
\paragraph{Models:} We evaluate seven RAG configurations: zero-shot prompting (without retrieval), vector retriever, hybrid of vector and BM25~\cite{10.1561/1500000019} retrievers, Raptor~\cite{sarthi2024raptorrecursiveabstractiveprocessing}, and three graph-based methods: GraphRAG (GRAG, ~\citeauthor{edge2025localglobalgraphrag}), HippoRAG 2 (HRAG,~\citeauthor{gutiérrez2025ragmemorynonparametriccontinual}), and Proposition-Graph RAG (PGRAG, ~\citeauthor{choubey2024distillsynthkgdistillingknowledgegraph}). We use \textit{text-embedding-3-large} from OpenAI embedding for vectorizing text and GPT-4o~\cite{openai2024gpt4technicalreport} for response generation. For agentic RAG, we adopt the ReAct~\cite{yao2023reactsynergizingreasoningacting,Liu_LlamaIndex_2022} framework, which combines a vector index for unstructured retrieval with eight structured search tools for querying employees, customers, GitHub PRs, and URLs (see Appendix~\ref{appendix:react} for details). 
We evaluate ReAct using proprietary and open-source LLMs, including GPT-4o, o4-mini, Gemini-2.5-Flash (gem$_{2.5f}$), DeepSeek-R1 (DS$_{r1}$, ~\citeauthor{deepseekai2025deepseekr1incentivizingreasoningcapability}), Llama-3.1-405B-Instruct (lam$_{405B}$), Llama-3.1-70B-Instruct (lam$_{70B}$, ~\citeauthor{grattafiori2024llama3herdmodels}), Llama-4-Maverick-17B-128E-Instruct (lam$_{4Mav}$, ~\citeauthor{meta_llama4_blog}), and DeepSeek-V3 (DS$_{v3}$, ~\citeauthor{deepseekai2025deepseekv3technicalreport}). We use the same set of eight LLMs to evaluate long-context reasoning. Additional implementation and hyperparameter details are provided in Appendix~\ref{appendix:baseline}.

\paragraph{Evaluation Metrics:} We follow standard RAG evaluation practices \cite{ragas2024} and use GPT-4o to assess answer quality. For content-based queries, we ask the LLM to rate each predicted answer on a Likert scale from 1 to 100 based on its factual accuracy and relevance, using the question and ground-truth answer as reference. For the other three query types, we use extraction-based evaluation. We prompt the LLM to extract specific information from the answer—such as URLs or pull request links for artifact-search, employee ID for people-search, and company name for customer-search. We then normalize the extracted text (e.g., by lowercasing and removing punctuation) and compute the F1 score based on exact matches with the ground-truth answer. For overall comparison, we also report the average score across all four query types, combining the Likert ratings for content queries and F1 scores for others.

\subsection{RAG Evaluation}

In~\cref{tab:result-rag}, we compare the performance of different RAG approaches on the {\datasetname} dataset in the full-RAG setting, where relevant evidence must be retrieved from the entire dataset given a question. 

The 0-shot baseline, which relies solely on the built-in knowledge of GPT-4o without any retrieval, performs very poorly. It achieves 0\% accuracy on people-, customer-, and artifact-search queries. This result highlights that the model cannot infer answers without access to context, even though the dataset itself was generated by GPT-4o. The questions require reasoning over specific enterprise artifacts, which are not part of the model’s internal knowledge.

\textbf{All standard RAG systems show limited performance on the {\datasetname} dataset.} Notably, advance methods like Raptor and GraphRAG perform much worse than the simple vector baseline, while HippoRAG-2 and proposition-graph RAG perform only slightly better. The basic Hybrid baseline achieves the highest average performance score of 20.61, but overall performance remains low.

\textbf{The ReAct agent leads to a significant improvement in performance over all standard RAG approaches.} Using GPT-4o as the underlying model, ReAct achieves an average performance score of 32.96, 12.35 points higher than the Hybrid baseline. This highlights the effectiveness of combining reasoning and retrieval in an interactive, agent-based framework. ReAct benefits from structured search tools that improve retrieval quality and enable effective multi-step reasoning, particularly for non-content queries that involve structured data.

\textbf{All RAG systems struggle to reliably identify unanswerable queries.} Among baseline methods, GraphRAG performs best, correctly identifying 63.41\% of unanswerable queries, while the hybrid method lags at 25.32\%. Although ReAct enhances reasoning over retrieved content, its effectiveness in detecting unanswerable queries varies significantly across LLMs—ranging from 63.66\% with Gemini-2.5-Flash to just 6.01\% with o4-mini. The 0-shot approach answers correctly most of the unanswerable queries since no context has been provided. These again highlight a potential tradeoff between answerable and unanswerable shown in \citep{peng2024unanswerability}, and risk of hallucinated answers when no valid evidence exists.

\begin{table}[]
\centering
\resizebox{\columnwidth}{!}{%
\begin{tabular}{lrrrrrr} \toprule
Models & {Cont.} & {Peop.} & {Cust.} & {Art.} & {Avg.} & {Unan.} \\ \midrule
\multicolumn{7}{c}{Standard RAG Techniques (GPT-4o)} \\ \midrule
0-shot & 18.19 & 0.0 & 0.0 & 0.0 & 4.55 & \textbf{88.70} \\
Vector & 30.61 & 16.05 & 0 & 20.42 & 16.77 & 28.76 \\
Hybrid & 34.87 & 18.54 & 0 & 29.02 & 20.61 & 25.32 \\
Raptor & 31.42 & 12.68 & 0 & 14.98 & 14.77 & 47.82 \\
GRAG & 28.27 & 9.46 & 0 & 3.49 & 10.31 & 63.41 \\
HRAG& \textbf{39.35} & 11.38 & 0 & 18.12 & 17.21 & 52.07 \\
PGRAG &  35.78 & 14.59 & 0 & 18.61 & 17.25 & 45.49 \\ \midrule
\multicolumn{7}{c}{Agentic RAG (ReAct)} \\ \midrule
GPT-4o &  32.22 & 23.45 & \textbf{41.35} & \textbf{34.81} & \textbf{32.96} & 23.03 \\
o4-mini & 35.25 & \textbf{27.19} & 24.54 & 26.94 & 28.48 & 6.01 \\
DS$_{r1}$ & 32.44 & 18.73 & 6.16 & 18.47 & 18.95 & 34.33 \\
lam$_{405B}$ & 27.84 & 18.23 & 33.17 & 22.03 & 25.32 & 40.06 \\
lam$_{70B}$ & 25.44 & 13.71 & 33.10 & 18.70 & 22.74 & 41.06 \\ 
lam$_{4Mav}$ & 33.01 & 22.74 & 30.18 & 30.16 & 29.02 & 57.22 \\ 
DS$_{v3}$ & 36.17 & 26.62 & 5.87 & 27.42 & 24.02 & 23.18 \\ 
gem$_{2.5f}$ & 31.54 & 23.32 & 21.22 & 25.55 & 25.41 & 63.66 \\
\bottomrule
\end{tabular}
}
\caption{Evaluation results of standard RAG techniques (using GPT-4o) and the React Agentic RAG approach using various LLMs on the {\datasetname} dataset.
}
\label{tab:result-rag}
\end{table}

\begin{table*}[]
\centering
\resizebox{0.9\linewidth}{!}{%
\begin{tabular}{l|rrrrrr|rrrrrr} \hline
Models & {Cont.} & {Peop.} & {Cust.} & {Art.} & {Avg.} & {Unan.} & {Cont.} & {Peop.} & {Cust.} & {Art.} & {Avg.} & {Unan.} \\ \hline
& \multicolumn{6}{c|}{Product-Specific Long Context} & \multicolumn{6}{c}{Product-specific RAG (ReAct Agent)} \\ \hline
gem$_{2.5f}$ & \textbf{82.34} & \textbf{68.80} & \textbf{79.07} & \textbf{75.97} & \textbf{76.55} & \textbf{60.80} & 42.39 & \textbf{45.86} & 39.01 & \textbf{40.16} & 41.86 & \textbf{57.22} \\
GPT-4o & 52.58 & 37.15 & 19.11 & 46.53 & 38.84 & 59.94 & 42.71 & 40.49 & \textbf{53.41} & 37.04 & \textbf{43.41} & 29.33 \\
DS$_{r1}$ &  69.98 & 50.25 & 49.37 & 65.05 & 58.66 & 38.91 & \textbf{46.50} & 41.57 & 21.53 & 29.63 & 34.81 & 36.48 \\
lam$_{405B}$ & 29.96 & 17.80 & 6.08 & 18.97 & 18.20 & 52.79 & 38.11 & 33.95 & 43.45 & 25.55 & 35.27 & 34.76 \\
\hline
\end{tabular}
}
\caption{Evaluation results of different LLMs in long-context setting and product-specific RAG settings. In the long-context setting (left half), all product-specific data is concatenated and provided directly as context, without retrieval. In the product-specific RAG setting (right half), the ReAct agent retrieves only from product-specific documents. Full results for all LLMs ar eprovided in Appendix \Cref{tab:result-long-context-full}.}
\label{tab:result-long-context}
\end{table*}

\subsection{Long Context Evaluation}
In~\cref{tab:result-long-context}, we evaluate various LLMs with 131K context length for a long-context setting, where we slightly relax the task difficulty by providing only the relevant product-specific data as input, removing information related to other products. This reduces the input tokens to 3.33\%
compared to the original {\datasetname} setting. To compare this against retrieval-based reasoning over the same context, we also evaluate a RAG setup in which the ReAct agent retrieves only from the same context.

\textbf{LLMs struggle with long-context reasoning on \datasetname.}
Gemini-2.5-Flash performs the best, achieving the highest average score of 76.55. Other recent high-performing LLMs fall significantly behind, with the second-best, DeepSeek-R1, scoring 17.89 points lower on average. However, even Gemini-2.5-Flash shows a 23.45 points performance gap, indicating that long-context reasoning remains a challenging problem. Older and open-source models like the LLaMA-3.1-Instruct series perform considerably worse in our long-context setting.

\textbf{Retrieval Limits Performance for Strong LLMs in \datasetname.}
Recent LLMs like Gemini-2.5-Flash (76.55 vs. 41.86) and DeepSeek-R1 (58.66 vs. 34.81) perform significantly better in the long-context setting than in RAG, even when retrieval is restricted to product-specific documents. This gap highlights retrieval as a key bottleneck, limiting the effectiveness of even reasoning-capable models. In contrast, models with weaker reasoning, such as LLaMA-3.1 variants, show gains in the product-specific RAG setting—likely due to reduced input length—but their performance drops when retrieval spans the full dataset without product boundaries (\Cref{tab:result-rag}).
These results underscore the challenge of retrieving relevant context in {\datasetname} and the need for retrieval systems that support deep search.


\begin{table}[]
\small
\centering
\begin{tabular}{lrrrrr} \toprule
Models & {Cont.} & {Peop.} & {Cust.} & {Art.} & {Avg.} \\ \midrule
GPT-4o & 61.72 & 52.99 & 63.28 & 68.94 & 61.73 \\
gem$_{2.5f}$ & 88.67 & \textbf{71.59} & \textbf{86.42} & 96.34 & \textbf{85.76} \\
DS$_{r1}$ & 84.66 & 63.84 & 68.85 & 90.31 & 76.92 \\
lam$_{405B}$ & 79.95 & 60.55 & 59.74 & 80.03 & 70.07 \\
\bottomrule
\end{tabular}
\caption{Evaluation results of different LLMs when using only the oracle evidence linked to each question (as identified during data synthesis) as context. Full results for all LLMs are provided in Appendix~\Cref{tab:result-oracle-full}.} \label{tab:result-citations}
\end{table}

\subsection{Oracle Evaluation}
\Cref{tab:result-citations} shows the performance of various LLMs in an oracle setting, where only the ground-truth evidence linked to each question, identified during data synthesis, is provided as context. This setup removes the variability introduced by retrieval (as in RAG) and unrelated or noisy context (as in long-context settings), enabling a controlled comparison of each model’s reasoning capabilities when given only the exact supporting information. This reduces the number of input tokens to just 0.213\% of the original {\datasetname} data.

\textbf{LLMs struggle to replicate human-designed reasoning on {\datasetname}.} In an oracle setting, overall performance improves, particularly for weaker models such as the LLaMA-3.1 variants. However, no model achieves perfect accuracy, even with full access to the relevant context. This highlights a key limitation of current LLMs: while the answers were generated using the same evidence, they were constructed through human-defined reasoning steps. In contrast, LLMs rely on their own inference processes and parametric knowledge of enterprise processes, which proves insufficient for consistently arriving at the correct answers. 


\section{Human Analysis}
\subsection{Common Errors in Oracle Setting}
We analyze model failures under oracle settings to better understand the error patterns of Gemini-2.5-Flash and DeepSeek-R1.  Among 815 answerable questions, Gemini scores zero on 71, while DeepSeek-R1 does so for 108. We examine 50 such questions per model. Our LLM-based evaluator performs consistently with human judgements in identifying incorrectly answered questions. The observed errors fall broadly into three categories.

\paragraph{Unanswered questions:} DeepSeek-R1 fails to answer 15 questions, either by using all output tokens for internal reasoning without producing a final answer or by explicitly stating that it cannot answer based on the provided context. Gemini exhibits similar behavior on 2 questions.

\paragraph{Incorrect reasoning:} DeepSeek-R1 answers 24 questions with flawed reasoning, while Gemini does so for 32 questions. These errors primarily occurred in questions related to the previous release of a product.  
A common mistake made by both models is inferring that members of the current team—who are merely discussing artifacts from the prior release—were also contributors to that earlier work.
However, correct reasoning requires identifying individuals who were directly involved in creating those artifacts, such as the author of the product requirements document or the QA specialist who reported bugs during that release.

\paragraph{Incomplete use of context} DeepSeek-R1 and Gemini fail to consider the full set of relevant evidence in 11 and 16 cases, respectively, resulting in incorrect answers. This issue frequently occurs in analytical questions. For example, to answer “Which customer has the most active bugs?”, a model must gather all reported bugs from sources like Slack messages and meeting transcripts, remove those with resolution PRs, and include bugs that were previously resolved or dismissed but later reopened. Instead, both models focus only on reopened bugs and ignore unresolved ones, which results in incorrect answers.

\subsection{ReAct Trajectories}
We manually analyze how the GPT-4o-based ReAct agent uses tools across 50 questions from Table~\ref{tab:result-rag} results. In 21 cases, it relies only on unstructured search. In 24 cases, it uses two tools, while 4 cases involve three tools, and just 1 case requires four. \Cref{tab:react-trajectories} shows the full breakdown by question type.

Most tool-use sequences are short and shallow. The agent always invokes unstructured search—often by itself or in combination with structured tools when employee or customer metadata is needed. It tends to rely on the first retrieved result rather than performing deeper or iterative searches. For instance, out of 10 artifact-related questions, the agent uses only unstructured search in 5 cases and fails to invoke structured tools like PR or URL search, answering based on surface-level mentions in the text.
We also observe 8 cases of unnecessary tool usage, such as resolving an employee name that isn’t needed to answer the question. In 4 other cases, the agent uses tools incorrectly—for example, querying with an employee ID when the tool expects a name.


\section{Conclusion}
We introduce {\datasetname}, a new benchmark for evaluating RAG systems in realistic enterprise settings. Unlike previous benchmarks that focus on shallow reasoning over loosely connected documents, {\datasetname} requires deep search across diverse and naturally connected data sources such as Slack messages, GitHub pull requests, meeting transcripts and chats, and internal documents. Our experiments show that even the most advanced agentic RAG systems struggle on {\datasetname}, with retrieval emerging as the primary limitation. While long-context language models perform better when given all product-specific data, they still fall short of consistently following the detailed reasoning steps needed to answer questions accurately. These results highlight the need for improved retrieval and reasoning methods to support realistic and complex enterprise RAG tasks.

\section*{Limitations}
Constructing {\datasetname} requires significant human effort—designing realistic workflows, connecting related documents, and writing good questions and answers all require manual work. This makes it harder to scale compared to fully synthetic benchmarks that can be generated automatically. Also, even though {\datasetname} is designed to reflect real enterprise workflows, it only covers a small part of how organizations actually work. Therefore, RAG or deep search systems evaluated on {\datasetname} should not be explicitly tuned to the specific workflows or distractor patterns used to generate the data, as that would reduce the generalizability of results.
Lastly, {\datasetname} focuses only on software and product development. Using it in other areas like healthcare or finance would require input from domain experts to create accurate and useful workflows.

\bibliography{custom}

\appendix

\section{Experiments}
\subsection{Baseline RAG Configurations} \label{appendix:baseline}
We use the same retrieval pool for all baselines methods, which includes Slack messages, meeting transcripts and chats, internal documents, GitHub PR and URL metadata, and employee and customer-related metadata. The Vector and Hybrid baselines are implemented using LlamaIndex~\cite{Liu_LlamaIndex_2022}. The Vector baseline performs dense retrieval over the unified index. The Hybrid baseline combines dense (vector) and sparse (BM25) retrieval by retrieving the top-$k$ results from each and taking their union.

GRAG is implemented using the official codebase from \url{https://github.com/microsoft/graphrag}, with the \texttt{drift} mode enabled. HRAG uses the official HippoRAG implementation from \url{https://github.com/OSU-NLP-Group/HippoRAG}. RAPTOR is implemented using the standard LlamaIndex pack available at \url{https://github.com/run-llama/llama_index/tree/main/llama-index-packs/llama-index-packs-raptor}. PGRAG uses the proposition graph retriever proposed by~\citet{choubey2024distillsynthkgdistillingknowledgegraph}.

We set $k=20$ for all methods. We use GPT-4o for response generation and knowledge graph construction for graph-based retrievers.

\subsection{ReAcT} \label{appendix:react}
We use the ReAct agent implementation from LlamaIndex~\cite{Liu_LlamaIndex_2022} for retrieval using both unstructured and structured tools. For unstructured search, we use a single hybrid retriever with top-$k=20$, applied over the union of Slack messages, meeting transcripts, meeting chats, and internal documents. For structured search, we define eight tools designed to extract employee, customer, PR, or URL-related information:
\begin{itemize}[leftmargin=*]
\item \textbf{Employee Role $\leftrightarrow$ ID Search:} Enables bi-directional mapping between an employee role and employee IDs. Given a role (e.g., \textit{Software Engineer}), the tool returns the IDs of all employees with that role. Conversely, it can map an employee ID back to their role.
\item \textbf{Employee Name $\leftrightarrow$ ID Search:} Enables bi-directional mapping between employee names and their IDs. Slack messages and meeting transcripts often mention employees by name, so the agent uses this tool to resolve employee references in unstructured text.
\item \textbf{Company ID $\leftrightarrow$ Name Search:} Provides mapping between internal company identifiers and their full names, helping resolve customer references in unstructured text.
\item \textbf{PR Link $\rightarrow$ Metadata Search:} Retrieves structured metadata (e.g., title, author, reviewers, status) given a Github pull request link.
\item \textbf{URL Link $\rightarrow$ Metadata Search:} Returns meta data for internal document or webpage URLs.
\end{itemize}

Unless otherwise specified, the agent performs a single run in which it can invoke any combination of structured and unstructured search tools to gather relevant information and answer the query. Each run is limited to a maximum of 20 tool calls due to a context length limit of 131K tokens. If the agent’s response is unsatisfactory—as determined by an LLM-based quality check—we fall back to using the hybrid retriever.

\subsection{Additional Results}
\paragraph{Effect of Attempt Budget on ReAct Performance:}
We evaluate the ReAct agent by allowing it to make up to 1, 5, 10, or 20 reasoning attempts per question. In each attempt, the agent completes a full reasoning trajectory, and an LLM-based judge determibes whether it has produced a valid answer. If the agent answers a question at any point, we stop further attempts. If it fails to provide an answer within the allowed limit, we fall back to the hybrid retriever. As shown in Table~\ref{tab:result-rag-ablation}, increasing the maximum number of attempts consistently improves performance across all query types, with the most substantial gains in People, Customer and Artifact search tasks. The average performance score rises from 32.96 with a single attempt to 37.29 with 20 attempts, while the percentage of unanswered questions drops from 23.03\% to 19.17\%. These results demonstrate a trade-off: more attempts enhance accuracy but also increase the risk of hallucinated answers.

\begin{table}[]
\centering
\resizebox{\columnwidth}{!}{%
\begin{tabular}{lrrrrrr} \toprule
\#Attempt & {Cont.} & {Peop.} & {Cust.} & {Art.} & {Avg.} & {Unan.} \\ \midrule
1 & 32.22 & 23.45 & 41.35 & 34.81 & 32.96 & \textbf{23.03} \\
5 & 34.05 & 28.26 & 45.33 & 39.32 & 36.74 & 20.60 \\
10 &  \textbf{34.43} & 28.19 & 45.60 & 40.34 & 37.14 & 20.03 \\
20 & 33.21 & \textbf{28.83} & \textbf{46.41} & \textbf{40.71} & \textbf{37.29} & 19.17 \\
\bottomrule
\end{tabular}
}
\caption{Performance of the ReAct agent based on GPT-4o under different attempt budgets.}
\label{tab:result-rag-ablation}
\end{table}

\begin{table*}[]
\centering
\resizebox{\linewidth}{!}{%
\begin{tabular}{l|rrrrrr|rrrrrr} \hline
Models & {Cont.} & {Peop.} & {Cust.} & {Art.} & {Avg.} & {Unan.} & {Cont.} & {Peop.} & {Cust.} & {Art.} & {Avg.} & {Unan.} \\ \hline
& \multicolumn{6}{c|}{Product-Specific Long Context} & \multicolumn{6}{c}{Product-specific RAG (ReAct Agent)} \\ \hline
gem$_{2.5f}$ & \textbf{82.34} & \textbf{68.80} & \textbf{79.07} & \textbf{75.97} & \textbf{76.55} & \textbf{60.80} & 42.39 & \textbf{45.86} & 39.01 & \textbf{40.16} & 41.86 & \textbf{57.22} \\
GPT-4o & 52.58 & 37.15 & 19.11 & 46.53 & 38.84 & 59.94 & 42.71 & 40.49 & \textbf{53.41} & 37.04 & \textbf{43.41} & 29.33 \\
o4-mini & 68.05 & 46.33 & 56.04 & 60.73 & 57.79 & 32.90 & 42.12 & 42.73 & 28.35 & 35.58 & 37.20 & 6.58 \\
DS$_{r1}$ &  69.98 & 50.25 & 49.37 & 65.05 & 58.66 & 38.91 & \textbf{46.50} & 41.57 & 21.53 & 29.63 & 34.81 & 36.48 \\
lam$_{405B}$ & 29.96 & 17.80 & 6.08 & 18.97 & 18.20 & 52.79 & 38.11 & 33.95 & 43.45 & 25.55 & 35.27 & 34.76 \\
lam$_{70B}$ & 37.07 & 21.51 & 5.10 & 19.82 & 20.88 & 43.63 & 36.62 & 31.59 & 44.54 & 19.06 & 32.95 & 31.90\\ 
lam$_{4Mav}$ & 60.91 & 44.54 & 50.70 & 44.87 & 50.26 & 44.64 & 43.35 & 43.64 & 39.28 & 35.04 & 40.33 & 52.65 \\ 
DS$_{v3}$ & 63.66 & 47.26 & 48.27 & 56.99 & 54.05 & 18.31 & 46.09 & 44.78 & 39.02 & 31.44 & 40.33 & 17.10 \\ 
 \hline
\end{tabular}
}
\caption{Evaluation results of different LLMs in long-context setting and product-specific RAG settings. In the long-context setting (left half), all product-specific data is concatenated and provided directly as context, without retrieval. In the product-specific RAG setting (right half), the ReAct agent retrieves only from product-specific documents.}
\label{tab:result-long-context-full}
\end{table*}

\begin{table}[]
\small
\centering
\begin{tabular}{lrrrrr} \toprule
Models & {Cont.} & {Peop.} & {Cust.} & {Art.} & {Avg.} \\ \midrule
GPT-4o & 61.72 & 52.99 & 63.28 & 68.94 & 61.73 \\
o4-mini & \textbf{89.47} & 64.24 & 75.07 & 91.23 & 80.0 \\
gem$_{2.5f}$ & 88.67 & \textbf{71.59} & \textbf{86.42} & 96.34 & \textbf{85.76} \\
DS$_{r1}$ & 84.66 & 63.84 & 68.85 & 90.31 & 76.92 \\
lam$_{405B}$ & 79.95 & 60.55 & 59.74 & 80.03 & 70.07 \\
lam$_{70B}$ & 71.89 & 56.75 & 54.01 & 75.43 & 64.52 \\ 
lam$_{4Mav}$ & 84.35 & 67.02 & 73.07 & \textbf{96.55} & 80.25  \\ 
DS$_{v3}$ & 82.39 & 64.52 & 54.54 & 91.52 & 73.24 \\ \bottomrule
\end{tabular}
\caption{Evaluation results of all LLMs when using only the oracle evidence linked to each question (as identified during data synthesis) as context.}
\label{tab:result-oracle-full}
\end{table}

\begin{table*}[]
\centering
\small
\resizebox{0.95\linewidth}{!}{%
\begin{tabular}{p{12cm}cccc}
\toprule
\textbf{Trajectory} & \textbf{Cont.} & \textbf{Peop.} & \textbf{Cust.} & \textbf{Art.} \\
\midrule
unstructured\_search & 15 & 1 & -- & 5 \\
unstructured\_search + \textbf{employee\_name\_to\_ID} + \textit{employee\_ID\_to\_name$^*$} & -- & 1 & -- & -- \\
unstructured\_search + \textbf{customer\_name\_to\_ID} & -- & -- & 2 & -- \\
unstructured\_search + \textit{employee\_ID\_to\_name$^*$} & -- & 5 & -- & -- \\
unstructured\_search + pr\_search & -- & 1 & -- & 4 \\
unstructured\_search + pr\_search + \textit{employee\_ID\_to\_name$^*$} & -- & 1 & -- & -- \\
unstructured\_search + unstructured\_search & 1 & 1 & -- & 1 \\
employee\_role\_to\_ID + unstructured\_search & -- & 2 & -- & -- \\
unstructured\_search + \textbf{customer\_name\_to\_ID} + unstructured\_search & 1 & -- & -- & -- \\
unstructured\_search + customer\_ID\_to\_name & -- & -- & 6 & -- \\
unstructured\_search + employee\_ID\_to\_role & -- & 1 & -- & -- \\
unstructured\_search + unstructured\_search + \textit{employee\_ID\_to\_name$^*$} & -- & 1 & -- & -- \\
unstructured\_search + employee\_role\_to\_ID + unstructured\_search + unstructured\_search & -- & 1 & -- & -- \\
\bottomrule
\end{tabular}
}
\caption{Count of ReAct trajectories by question type. Each row represents a distinct trajectory executed by the ReAct agent, and the columns indicate the number of questions of each type that followed the corresponding trajectory. “--” indicates no questions of that type were associated with the trajectory. \textbf{Bolded} tool names within trajectories denote that the tool was invoked with incorrect parameters. \textit{Italicized$^*$} tool names indicate that the tool was called unnecessarily, i.e., its output was not required for answering the question.}
\label{tab:react-trajectories}
\end{table*}


\section{Dataset}
\subsection{Example Data Schema Objects in {\datasetname}} \label{app:schema-objects}
\Cref{employee-schema} shows an example team with roles such as engineers and product managers. \Cref{product-schema} shows sample product metadata used during data generation; however, this metadata is not typically available in real enterprise environments and should not be used to evaluate system performance. Similarly, \Cref{customer-schema} illustrates an example customer profile, including associated products. Since product-customer links are rarely stated explicitly in real enterprise data, this information is also excluded from performance evaluation. \Cref{slack-schema}, \Cref{meeting-transcript-schema}, and \Cref{meeting-chat-schema} illustrate Slack messages, meeting transcripts, and chats. \Cref{dot-schema}, \Cref{pr-schema}, and \Cref{url-schema} show examples of documents, GitHub PRs, and URL metadata.

\subsection{Distractors for Enterprise Noise Modeling} \label{app:distractor}
To increase the complexity of the {\datasetname} data synthesis pipeline and better mimic real-world enterprise noise, we introduce several distractors—elements that complicate the reasoning process without contributing to the correct answer. 

\begin{itemize}[leftmargin=*]
\item \textbf{Temporal Overlap:} Teams simultaneously discuss related topics (e.g., competitor product analysis) while producing planning documents, requiring models to topically disentangle mixed discussions.
\item \textbf{Product Name Change:} A product may be renamed during the planning or development stage. After the change, all future documents and messages use the new name, so models must link both names to the same product.
\item \textbf{Multiple Planning Teams:} More than one team initiates product planning, but only one continues to development and support stages. Queries may involve distinguishing which team authored artifacts that persist through all stages.
\item \textbf{Cross-Product References in Development:} Teams discuss similar features from competitors or open-source libraries and share corresponding GitHub PRs during development-stage meetings and Slack threads.
\item \textbf{Cross-Product References in Support:} Teams bring up bugs found in other projects—like competitor software or open-source tools—that are similar to their own issues in stage three workflows.
\item \textbf{Public External Links:} Teams often share public content—such as blogs, demo videos, or news articles—which may seem topically related but are not about the current product.
\item \textbf{Legacy Feedback Mentions:} Some planning discussions mention customer or QA feedback from previous product versions, which were created by different teams, making it harder to track who was involved.
\item \textbf{Competitor Planning Documents:} Teams sometimes review or comment on planning documents for other (competitor) products, adding cross-product entanglement into planning-stage reasoning.
\end{itemize}


\begin{table*}[ht]
\small
\centering
\begin{tabular}{>{\centering\arraybackslash}p{0.04\linewidth}
     >{\centering\arraybackslash}p{0.03\linewidth}
     p{0.82\linewidth}}
\toprule
\textbf{Stage} & \textbf{ID} & \multicolumn{1}{p{0.82\linewidth}}{\centering \textbf{Description}} \\ \midrule
1 & 1 & The Marketing Research Analyst, Product Manager, and Architect each draft their respective documents. All documents undergo review and revision through Slack discussions. Teams also discuss strengths and weaknesses of competitor products during this phase. \\ \midrule
1 & 2 & The same documents are created in a two-stage review process: initial feedback via Slack and final feedback through meetings. Teams also review competitor documents during this process. \\ \midrule
1 & 3 & The new team reviews key documents from the previous release and discusses customer feedback and QA issues. Based on this, they update the same set of documents for the next release. \\ \midrule
2 & 1 & The workflow begins with a team meeting to discuss new features for the upcoming release. In regular follow-up meetings, the team assigns implementation tasks to engineers and tracks progress. Engineers create pull requests (PRs), which are reviewed and either approved or rejected and reassigned through Slack discussions or meetings. Throughout the process, the team also shares and discusses PRs for similar features in open-source libraries to inform their development. \\ \midrule
2 & 2 & As a continuation of the first workflow, some approved or merged PRs are later reverted due to changes in product direction, security concerns, or other factors. \\ \midrule
2 & 3 & As a continuation of the first workflow, some approved PRs are put on hold due to changes in product direction, security concerns, or other factors, and are merged later once the issues are resolved. In some cases, previously merged PRs are also reverted for similar reasons. \\ \midrule
3 & 1 & The team resolves customer-reported bugs, which come from multiple customers and may have similar impact. Bugs are discussed and assigned to engineers through Slack. Engineers then fix the issues and submit pull requests (PRs) for review. Throughout the process, the team also discusses similar bugs and fixes in open-source projects to inform their approach. \\ \midrule
3 & 2 & As a continuation of the first workflow, engineers investigate assigned customer-reported bugs. In some cases, they determine that the reported behavior is expected and not a real issue. For valid bugs, they implement fixes and submit pull requests (PRs) for review. \\ \midrule
3 & 3 & As a continuation of the second workflow, some dismissed or resolved bugs are re-reported by customers. The team revisits these cases, re-evaluates the original assessments or fixes, and makes updates if needed. Engineers submit revised pull requests (PRs) where necessary to address the recurring issues. \\ \bottomrule
\end{tabular}
\caption{Overview of nine enterprise workflows, three per stage, used for synthesizing {\datasetname} dataset.} 
\label{app-tab:serb-wfs}
\end{table*}

\begin{figure}[]
\tiny
\begin{verbatim}
  {
    "employee_id": "eid_9b023657",
    "name": "Hannah Taylor",
    "role": "VP of Engineering",
    "location": "San Francisco",
    "engineering_leads": [
      {
        "employee_id": "eid_e96d2f38",
        "name": "David Williams",
        "role": "Engineering Lead",
        "location": "Remote",
        "engineers": [
          {
            "employee_id": "eid_234b3360",
            "name": "Ian Davis",
            "role": "Software Engineer",
            "location": "Remote",
            "org": "slack"
          },..
        ],
        "qa_specialists": [
          {
            "employee_id": "eid_df392037",
            "name": "George Taylor",
            "role": "QA Specialist",
            "location": "Remote",
            "org": "slack"
          },..
        ],,
        "org": "slack"
      }..
    ],
    "product_managers": [
      {
        "employee_id": "eid_03a183c9",
        "name": "Fiona Brown",
        "role": "Product Manager",
        "location": "Seattle",
        "org": "slack"
      },..
    ],
    "tech_architects": [
      {
        "employee_id": "eid_816aea15",
        "name": "Alice Taylor",
        "role": "Technical Architect",
        "location": "San Francisco",
        "org": "slack"
      }
    ],
    "ux_researchers": [
      {
        "employee_id": "eid_839bc5eb",
        "name": "Julia Jones",
        "role": "UX Researcher",
        "location": "Seattle",
        "org": "slack"
      }
    ],
    "marketing_research_analysts": [
      {
        "employee_id": "eid_bb22e59b",
        "name": "Alice Martinez",
        "role": "Marketing Research Analyst",
        "location": "Remote",
        "org": "slack"
      }
    ],
    "chief_product_officers": [
      {
        "employee_id": "eid_ce2f3276",
        "name": "Bob Williams",
        "role": "Chief Product Officer",
        "location": "New York",
        "org": "slack"
      }
    ],
    "marketing_managers": [
      {
        "employee_id": "eid_f1c8f4a5",
        "name": "Charlie Martinez",
        "role": "Marketing Manager",
        "location": "Berlin",
        "org": "slack"
      }
    ],
    "org": "slack"
  }     
\end{verbatim}
\caption{An example Team.} 
\label{employee-schema}
\end{figure}

\begin{figure}[]
\tiny
\begin{verbatim}
{
  "name": "MonitorForce",
  "product": "MuleSoft AI Monitoring",
  "alias": "torAIX",
  "description": "AI-driven monitoring tool that provides real-time 
  insights and proactive alerts on API performance and health.",
  "distractor": "MuleSoftAIMonitoring",
  "old_name": "torProX",
  "competitor": {
    "name": "New Relic AI Monitoring",
    "description": "monitors distributed systems and APIs in real time, 
    delivering alerts on performance issues."
  }
}
\end{verbatim}
\caption{An example product-related metadata.} 
\label{product-schema}
\end{figure}

\begin{figure}[]
\tiny
\begin{verbatim}
{
  "name": "Frank Lewis",
  "role": "Product Manager",
  "company": "BlueWave",
  "products": [
    "SupportForce",
    "VizForce",
    "WorkflowForce",
    "SentimentForce",
    "ForecastForce",
    "ContentForce"
  ],
  "id": "CUST-0001"
}
\end{verbatim}
\caption{An example customer profile.} 
\label{customer-schema}
\end{figure}

\begin{figure}[]
\tiny
\begin{verbatim}
{
  "Channel": {
  "name": "planning-SecurityForce-PM", 
  "channelID": "ch-ix-pm-16b727"
  }, 
  "Message": {
    "User": {
      "userId": "eid_c8ebc4b0", 
      "timestamp": "2026-04-28T10:21:00", 
      "text": "Hi team, I've shared the <https://sf-internal.slack.com/
      archives/docs/rityaix_product_vision_document|Product Vision Doc
      ument> for rityAIX. Let's discuss any feedback or suggestions you 
      might have to refine it further. Looking forward to your thoughts", 
      "utterranceID": "20260427-0-a635b"
    }, 
    "Reactions": []
  }, 
  "ThreadReplies": []
}
\end{verbatim}
\caption{An example slack message.} 
\label{slack-schema}
\end{figure}

\begin{figure}[]
\tiny
\begin{verbatim}
{
  "name": "product_dev_WorkFlowGenie_5", 
  "text": "Attendees\nGeorge Garcia, David Garcia, Ian Davis, David Brown, 
          Julia Smith, Alice Taylor, Julia Brown, Charlie Davis, 
          Charlie Martinez, Fiona Martinez, Bob Miller, Julia Garcia, 
          Alice Taylor, Fiona Martinez, Bob Garcia, Charlie Smith, 
          Alice Johnson\n
          Transcript\n
          George Garcia: Alright team, let's kick off this sprint review. 
          First, let's go over the completed PRs. Julia, can you...
          ...
          George Garcia: Alright then, let's get to work. Thanks", 
  "timestamp": "2026-04-29T06:27:00", 
  "participants": [
    "eid_95f6d01c", "eid_f4f58faa", "eid_71c0d545", "eid_1f678d18", 
    "eid_827a0ea9", "eid_2543da6a", "eid_c92d3e03", "eid_4812cbd8",
    "eid_55f29a0d", "eid_1e8695b6", "eid_d96bfd9b", "eid_136119e9", 
    "eid_18571957", "eid_e214d622", "eid_e3c15ff5", "eid_515ae627", 
    "eid_686130c8"
  ]
}
\end{verbatim}
\caption{An example meeting transcript.} 
\label{meeting-transcript-schema}
\end{figure}

\begin{figure}[]
\tiny
\begin{verbatim}
{
  "text": "2026-02-11T01:00:00\nHannah Taylor: https://sf-internal.
  slack.com/archives/docs/final_ectaix_product_vision_document", 
  "id": "ectAIX_planning_2_chat"
}
\end{verbatim}
\caption{An example meeting chat.} 
\label{meeting-chat-schema}
\end{figure}

\begin{figure}[]
\tiny
\begin{verbatim}
{
  "name": "new_shaix_system_design_document", 
  "link": "https://sf-internal.slack.com/archives/docs/new_shaix_s
           ystem_design_document", 
  "text": "Introduction: Salesforce AI Search is an advanced enterprise 
           search solution designed to optimize data retrieval within the 
           Salesforce ecosystem. By leveraging machine learning.....", 
  "timestamp": "2026-10-06T08:04:00", 
  "author": "eid_d0b6cb92", 
  "title": "System Design Document"
}
\end{verbatim}
\caption{An example document.} 
\label{dot-schema}
\end{figure}

\begin{figure}[]
\tiny
\begin{verbatim}
{
  "id": "github_com_LibreOffice_core_pull_1576", 
  "title": "Data Corruption in Calc Reports", 
  "summary": "Generated spreadsheets sometimes contain corrupted 
             summary data, misleading users about financial 
             calculations.", 
  "link": "https://github.com/LibreOffice/core/pull/1576", 
  "mergeable": true, 
  "merged": true, 
  "number": "1576", 
  "state": "closed", 
  "user": {
    "login": "EMP_361702750"
  }, 
  "created_at": "2025-03-12T00:00:00", 
  "reviews": [
    {
      "state": "APPROVED", 
      "user": {
        "login": "EMP_310841237"
      }, 
      "comment": "LGTM", 
      "submitted_at": "2025-03-13T01:46:00"
    }
  ]
}
\end{verbatim}
\caption{An example GitHub PR metadata.} 
\label{pr-schema}
\end{figure}

\begin{figure}[]
\tiny
\begin{verbatim}
{
  "id": "hbr_org_2023_02_the-ethical-implications-of-ai-in-sales", 
  "link": "https://hbr.org/2023/02/the-ethical-implications-of-ai
           -in-sales",
  "description": "A Harvard Business Review article discussing the 
                 ethical implications of AI in sales and coaching."
}
\end{verbatim}
\caption{An example meeting URL metadata.} 
\label{url-schema}
\end{figure}

\begin{table*}[htbp]
\centering
\resizebox{0.95\linewidth}{!}{%
\small
\begin{tabular}{p{0.11\linewidth} p{0.89\linewidth}}
\toprule
\textbf{Query Type} & \textbf{Query Templates} \\
\midrule

\multirow{6}{*}{\textbf{Content}} 
& What are the unique features of \textit{\{product\}}'s competitor products? \\
& What are the weaknesses of \textit{\{product\}}'s competitor products? \\
& What are the changes suggested by \textit{\{employee role\}} to improve the \textit{\{document name\}} for \textit{\{product\}}? \\
& What issues were reported by customers for the previous release of \textit{\{product\}}? \\
& What issues were reported during QA testing for the previous release of \textit{\{product\}}? \\
& What strengths were highlighted by customers for the previous release of \textit{\{product\}}? \\
& Find features for \textit{\{product\}} that were discussed but not implemented? \\
& What new features about \textit{\{feature topic\}} are added to the \textit{\{product\}}? \\
& What features could not be added to \textit{\{product\}} due to \textit{\{issue type\}}? \\
& What features were temporarily delayed due to \textit{\{issue type\}} but eventually added to \textit{\{product\}}? \\
& Find bugs reported by \textit{\{customer company name\}} in \textit{\{product\}} that did not require any fixes? \\
& Find all unresolved issues reported by \textit{\{customer company name\}} in the \textit{\{product\}}? \\
\midrule

\multirow{7}{*}{\textbf{People}} 
& Find employee IDs of team members who provided insights on the strengths and weaknesses of \textit{\{product\}}'s competitor products. \\
& Find employee IDs of team members who shared demos of \textit{\{product\}}'s competitor products. \\
& Find employee IDs of the authors and key reviewers of the \textit{\{document name\}} for the \textit{\{product\}} product. \\
& Find employee IDs of \textit{\{employee role\}}s who worked on the previous release of \textit{\{product\}}. \\
& Find employee IDs of team members who shared demos of the previous version of \textit{\{product\}}. \\
& Find employee IDs of QA specialists who worked on the previous release of \textit{\{product\}}. \\
& Find employee IDs of team members who were responsible for features in \textit{\{product\}} that were discussed but not implemented. \\
& Find the employee ID of engineer with the highest number of approved feature development PRs in {product['name']}. \\
& Find the employee ID of engineer with the highest number of unapproved feature development PRs in {product['name']}. \\
& Find employee IDs of engineers who resolved the issues with \textit{\{issue area and impact\}} in \textit{\{product\}}. \\
& Find the employee ID of engineer who resolved the maximum number of customer bugs in \textit{\{product\}}. \\
& Find the employee ID of engineer who is assigned to the highest number of unresolved bugs in \textit{\{product\}}. \\
& Find the employee ID of engineer who authored maximum number of PRs in {product name} for bug fixes that are now reopened by customers. \\
& Find the employee ID of engineer who dismissed maximum bugs in \textit{\{product\}} that are now reopened by customers. \\
\midrule

\multirow{7}{*}{\textbf{Artifact}} 
& Find demo URLs shared by team members for \textit{\{product\}}'s competitor products. \\
& Find demo URLs shared by team members for the previous version of \textit{\{product\}}. \\
& Find links to PRs for implementing \textit{\{feature topic\}} in \textit{\{product\}} that were reverted due to \textit{\{issue type\}}. \\
& Find links to the approved PRs for implementing \textit{\{feature topic\}} in \textit{\{product\}} that need to be closed. \\
& Find links to the PRs for implementing \textit{\{feature topic\}} in \textit{\{product\}} that need to be merged. \\
& Find links to PRs for \textit{\{feature topic\}} in \textit{\{product\}} that were not approved. \\
& Find link to the approved PRs for implementing \textit{\{feature topic\}} in \textit{\{product\}}. \\
& Find PR links submitted by \textit{\{employee ID\}} to resolve bugs reported by \textit{\{customer company name\}} for \textit{\{product\}}. \\
\midrule

\multirow{7}{*}{\textbf{Customer}} 
& Find names of companies which reported issues with the previous release of \textit{\{product\}}. \\
& Find names of companies which highlighted strengths of the previous release of \textit{\{product\}}. \\
& Find names of companies which reported the issues with \textit{\{issue area and impact\}} in \textit{\{product\}}. \\
& Find the name of company which has the maximum number of resolved bugs in \textit{\{product\}}. \\
& Find the name of company which reported the maximum number of issues that didn’t need fixes in \textit{\{product\}}. \\
& Find the name of company which has the maximum number of reopened bugs in \textit{\{product\}}. \\
& Find the name of company which has the maximum number of active bugs in \textit{\{product\}}. \\
\bottomrule

\end{tabular}}
\caption{All 41 manually constructed query templates categorized into four enterprise search intents: content-based (12), people-search (14), artifact-search (8), and customer-search (7).}
\label{tab:question-templates}

\end{table*}

\end{document}